\documentclass[accepted]{uai2021} 

\usepackage[american]{babel}

\usepackage{natbib} 
\usepackage{mathtools} 
\usepackage{booktabs} 
\usepackage{tikz} 

\usepackage{amsfonts}

\DeclareMathOperator*{\argmin}{arg\,min}

\newcommand{\cc}[1]{{\cal #1}}

\newcommand{\expec}[1]{\mathbb{#1}}
\newtheorem{theorem}{Theorem}

\title{Differentiable Robust LQR Layers}

\author[1]{Ngo Anh Vien}
\author[2]{Gerhard Neumann}

\affil[1]{Bosch Center for Artificial Intelligence, Germany
}
\affil[2]{Karlsruhe Institute of Technology, Germany}

\begin{document}
\maketitle

\begin{abstract}
  This paper proposes a differentiable robust LQR layer for reinforcement learning and imitation learning under model uncertainty and stochastic dynamics. The robust LQR layer can exploit the advantages of robust optimal control and model-free learning. It provides a new type of inductive bias for stochasticity and uncertainty modeling in control systems. In particular, we propose an efficient way to differentiate through a robust LQR optimization program by rewriting it as a convex program (i.e. semi-definite program) of the worst-case cost. Based on recent work on using convex optimization inside neural network layers, we develop a fully differentiable layer for optimizing this worst-case cost, i.e. we compute the derivative of a performance measure w.r.t the model's unknown parameters, model uncertainty and stochasticity parameters. We demonstrate the proposed method on imitation learning and approximate dynamic programming on stochastic and uncertain domains. The experiment results show that the proposed method can optimize robust policies under uncertain situations, and are able to achieve a significantly better performance than existing methods that do not model uncertainty directly.
\end{abstract}

\section{Introduction}

Combining model-free and model-based reinforcement learning (RL) has recently received much attentions by researchers in the community. While model-free approaches can be easy for training, they suffer from poor sample efficiency and weak generalization and being too task-specific. Therefore they might have limited applicability to real-world physical systems, where long-time running would lead to being unsafe or to break the system. In contrast, model-based approaches are known to be sample-efficient, however when trained on a physical system they often suffer from a similar safety and robustness issues. These issues can possibly mislead the system to breaking states \citep{amodei2016concrete}. Model learning and policy optimization are sometimes treated as two separate stages, which is a common approach in dual control, i.e. learning and control \citep{feldbaum1960dual}. However, they can not fully exploit the advantages of domain knowledge in an end-to-end fashion, i.e. \emph{inductive bias of optimal control}. Injecting inductive bias of optimal control in end-to-end RL algorithms has recently shown many great successes in terms of sample-efficiency and robustness \citep{tamar2016value,silver2017predictron,karkus2017qmdp,FarquharRIW18, HafnerLB020}. 

We follow work using inductive bias of optimal control in end-to-end RL \citep{AmosRSBK18}. While existing work has not yet included model uncertainty and dynamics stochasticity, modeling uncertainty and perturbation is considered very important for robot learning  \citep{mankowitz2019robust}. In this paper, we will propose a differentiable robust linear quadratic regulator (LQR) layer that introduces an \emph{inductive bias of robust optimal control}. We combine the advantages of model-based robust optimal control and model-free policy learning. Based on recent work on using convex optimization inside neural network layers \citep{agrawal2019differentiable}, our algorithm differentiates through a robust optimization algorithm, and hence becomes end-to-end trainable without the necessity of unrolling the planning procedure. Specifically, similar to a standard neural network (NN) layer the developed robust optimization layer computes an optimal solution of a robust optimization program as output. This layer can be differentiated through, and integrated into a NN as a differentiable module to achieve an end-to-end task network. The parameters of this layer are parameters used in the parameterization of the dynamics model, uncertainty set, and the distribution of the dynamics stochasticity. Different from existing approaches, our robust LQR layer is able to model stochasticity (stochastic dynamics models), and uncertainty (uncertainty set over models). We will show that we can differentiate through this robust layer by resorting to the Linear Matrix Inequality (LMI) formulation \citep{boyd1994linear}. In particular, we rewrite robust constraints as the worst-case cost, and derive an approximation to the objective of the robust control program. The resulting convex semi-definite program can then be differentiated using the differentiable conic program algorithm as introduced recently by Agrawal et. al. \citep{agrawal2019differentiating}.

We will show that on imitation learning under the presence of task uncertainty we can estimate both i) the model parameters, i.e. the cost and dynamics, and also ii) the model uncertainty, i.e. the stochasticity of the dynamics and the model parameter's uncertainty. The uncertainty might come from environment disturbances or from the expert's non-stationary model, i.e. where the environment dynamics or the expert's mental state change.



\section{Related Work}

{\bf Model-free and model-based} methods are two common approaches in RL. Model-free approaches are learning direct mappings from raw inputs to actions \citep{mnih-atari-2013,LillicrapHPHETS15,schulman2015trust}. Although these approaches might be easy to implement and can make training fast and easy, they have weak generalization ability, are not sample-efficient, and tend to be task-specific \citep{sun2019towards}. On the other hand, model-based approaches learn an approximate model of the environment dynamics from real interactions, and optimize an optimal policy using fictitious data generated from the learnt dynamics model. There have been efforts to combine the advantages of model-based RL with control theory, where optimal control is used both as domain knowledge embedded in model learning and as a solver for policy optimization, i.e. model predictive control (MPC), linear quadratic regulator (LQR), or differential dynamic programming (DDP) \citep{LevineK13,zhang2016learning,nagabandi2018neural,hafner2019learning}. 

{\bf Differentiable optimization layer}: There has been recent effort in using convex optimization inside neural network layers. Most existing model-based methods propose to integrate unrolling out planning steps into policy networks, i.e. a connection between two layers is defined as coming from a planning step. \citep{tamar2016value,karkus2017qmdp,silver2017predictron,oh2017value,FarquharRIW18}. However this rolling procedure makes forward and backward calculations become computationally expensive. There are recent approaches that can handle analytic differentiation of optimization, hence they have more efficient computation. In particular, Mensch et. al. \citep{mensch2018differentiable} propose a differentiable dynamics programming layer. \cite{abs-2101-09207} propose a differentiable trust region layer that can be integrated directly into TRPO and PPO \citep{schulman2015trust}. A concurrent work by \cite{abs-2011-08105} also proposes to integrate robust control as a custom convex-optimization-based projection layer for generic nonlinear control policy networks. Other work \citep{amos2017optnet,agrawal2019differentiable} incorporate general convex programs inside deep networks. However it is non-trivial to extend those approaches to incorporate a robust optimal control program within deep policy networks. 

{\bf Safe reinforcement learning} \citep{garcia2015comprehensive} is concerned with how safety should be addressed in \emph{learning to control} systems. Safety can be defined through risk-averse reward functions \citep{coraluppi1999risk} where high costs are at risk or at undesirable states. Similarly, safe exploration \citep{SchreiterNEBMT15,achiam2017constrained} and robust policy search \citep{mankowitz2019robust} guide RL agents to explore the state space while adhering to certain safety constraints while the policy is being optimized. In addition, robust Markov decisions processes (MDPs) \citep{iyengar2005robust} and robust optimal control \citep{vandenberghe2002robust,bemporad1999robust} consider policy optimization under the presence of uncertainty over the dynamics models and state knowledge. These approaches require an explicit definition of safety, e.g. a risk-averse reward function, or auxiliary cost constraint functions. Alternatively, robust dual control is concerned with simultaneous learning and control subject to uncertainty and stochastic dynamics \citep{abbasi2011regret,weiss2014robust,cohen2018online,DeanMMRT18,dean2019safely}. These algorithms are able to offer a differentiable objective for policy optimization and model estimation, however they cannot perform full end-to-end learning, i.e. the possibility of being integrated into a fully differentiable task network. 

\section{Background}
In this paper, we are concerned with the robust infinite-horizon linear-quadratic-Gaussian (LQG) control problem under ellipsoidal uncertainty,
\begin{equation}
\small
\begin{aligned}
  \pi^* &= \argmin_{u=\pi(x)} \sum_{t=0}^\infty  \left(x^{\top}_t Q x_t + u^{\top}_t R u_t\right), \\
   \text{s.t.} \quad & x_{t+1} =Ax_t + Bu_t + w_t \\ &  w_t \in {\cal N}(0,\sigma^2I_n) ,\quad x_0 \sim P_0,\\
  & [A,B]  \in \{A,B: \left(X^{\top} - \mu \right) ^{\top} D \left(X^{\top} - \mu \right) \le I\},
\end{aligned}
\label{problem}
\end{equation}
where $x_t \in R^n, u_t \in R^m, A \in R ^{n\times n},B \in R^{n\times m}$. $I_n, I$ are $n\times n$ and $(m+n)\times (m+n)$ identity matrices, respectively. We denote $X=[A,B]$ a concatenated vector of $A$ and $B$; $P_0$ denotes an initial state distribution. The random variable $w_t$ denotes the noise of the stochastic dynamics, which is assumed to follow a Gaussian distribution ${\cal N}(0,\sigma^2 I_n)$, where $\sigma \in R$. $Q \in R ^{n\times n}, R \in R ^{m\times m}$ are positive definite matrices which model the quadratic costs. In addition, we assume the model parameter uncertainty is described by an ellipsoidal uncertainty that is parameterized by a nominal model $\mu=[\bar A ,\bar B]$ and a symmetric positive definite matrix $D$.




\subsection{Differentiable LQR}
\label{difflqr}
Recently there has been significant effort in using differentiable programs as a neural network layer, e.g. optimization layers \citep{agrawal2019differentiable}, differentiable ODE solvers \citep{chen2018neural}, MPC-based policy layers \citep{AmosRSBK18}. Amos et.al. \citep{AmosRSBK18} propose an algorithm that can differentiate through a policy that is represented by a LQR and MPC. In particular, Amos et.al. \citep{AmosRSBK18} use a discrete-time finite-horizon LQR (with a formulation similar to \eqref{problem} with a finite time horizon and without uncertainty and stochasticity, i.e. without the last two inequality constraints) as a learnable module with trainable parameters $\theta=\{A,B,Q,R\}$. The differentiation is made through the fixed-point solution $\tau^* =\{x^*_t,u^*_t\}_{t=1:T}$ of the LQR problem. The fixed-point solution can be found using the iterative method (involving a forward and a backward recursion). The main challenge is to compute the derivative $\partial l/\partial \theta$ of a generic loss function $l(\tau^*)$ of $\tau^*$ w.r.t the parameters $\theta$ of the LQR-based policy. This derivative can be written as $\partial l/\partial \theta =\frac{\partial l}{\partial \tau^*}\frac{\partial \tau^*}{\partial \theta}$. Then the authors suggested to compute the derivatives and differentiation through the constrained convex quadratic LQR problem at the fixed point by applying the implicit function theorem \citep{dontchev2009implicit}. The implicit mapping between $\tau^*$ and $\theta$ is expressed as the zero of the partial derivative of the Lagrange function $\nabla_\tau {\cal L}(\tau^*,\lambda^*)=0$, where $\cal L$ is the Lagrange function of the constrained optimization problem Eq.~\ref{problem}, and can be written as follows
\begin{align*}
  {  \cal L}(\tau,\lambda) = \sum_{t=1}^T  \left(x^{\top}_t Q x_t + u^{\top}_t R u_t\right)  +\sum_{t=0}^{T-1} \lambda^{\top}_t (Ax_t + Bu_t -x_{t+1}),
\end{align*}
where $T$ is the length of the time horizon, and $\lambda =\{\lambda_t\}_{t=0}^{T-1}$ with $\lambda_t \in R^n$ being a Langrange multiplier at time $t$. 

This LQR control layer can also be extended to become a differentiable MPC-based control layer by differentiating the convex approximation at a fixed point of the iterative approximation procedure, i.e. an iterative linearization of the dynamics and second-order Taylor approximation of the cost. Finite-horizon LQR-based policies are open-loop, therefore stability is not guaranteed \citep{bitmead1991riccati}. This problem could be mitigated through the online open-loop MPC-based extension \citep{AmosRSBK18}. 

However, it is non-trivial to extend this approach to obtain a differentiable robust control layer that would require to model stochastic dynamics and/or model uncertainty. A \emph{differentiable robust control layer} with a robust optimal control inductive bias is expected to better model the underlying problem, hence will improve the robustness of learnt policies. For example, demonstration data can be generated by a stochastic process with an uncertain model, as shown in Eq.~\ref{problem}. In addition, both the forward and backward recursion used by Amos et. al. \citep{AmosRSBK18} based on the constrained convex quadratic program formulation will suffer from a poor convergence and high computation when the horizon $T$ increases.

\subsection{LQR and Linear Matrix Inequalities}
\label{nouncertainty}
In the case of deterministic dynamics and without uncertainty, the problem in Eq.~\ref{problem} becomes an infinite-horizon LQR control problem with a standard formulation as follows
\begin{equation}
\begin{aligned}
  \pi^* &= \argmin_{\pi} \sum_{t=0}^\infty  \left(x^{\top}_t Q x_t + u^{\top}_t R u_t\right) \\
  & \text{s.t.} \quad x_{t+1} =Ax_t + Bu_t, \quad x_0 = x_{\tt{init}},
\end{aligned}
\label{problem-lqr}
\end{equation}
Its optimal controller is state-feedback, $u_t= Kx_t$, with $K=-(B^{\top} PB + R)^{-1} B^{\top} P A$, where $P$ is a positive definite matrix and can be computed by solving the Algebraic Riccati Equation (ARE) \citep{camacho2013model},
\begin{equation}
\begin{aligned}
  P = A^{\top}PA + Q - A^{\top} P B(B^{\top} PB + R) B^{\top}PA .
\end{aligned}
\label{lqr}
\end{equation}
A solution for $P$ can be found using iterative methods \citep{hewer1971iterative} (backward recursion) or by a convex SDP formulation through the use of Linear Matrix Inequalities (LMI) \citep{boyd1994linear,balakrishnan2003semidefinite}. The SDP formulation can be done via by assuming the ARE to be a Lyapunov-inequality. The objective is to minimize the trace of $P$,
\begin{equation}
  \begin{aligned}
    \max_P & \quad \tt{trace}(P)  \\
    &    \text{s.t.} \left[\begin{array}{ccc} R + B^{\top}P B & B^{\top}P A & 0\\
        A^{\top }P B & A^{\top}P A+ Q -P & 0 \\
        0 & 0 & P
      \end{array} \right] \succeq 0,
  \end{aligned}
  \label{lmi}
\end{equation}
where $P$ is symmetric positive definite: $P\succeq 0, P=P^{\top}$. 

Stability of infinite-horizon LQR has been extensively studied in optimal control \citep{kalman1960contributions}. It has been shown via the Lyapunov analysis that LQR is robust to uncertainty in the model parameters (e.g. the $A$ and $B$ matrices) and to perturbations (e.g. the noise $w$) under certain conditions, i.e. the bounds on the uncertainty. In other words, as long as the uncertainty and perturbations are small enough the solution to the LMI constraint problem could exist and could stabilize the controller. We will show that such a limited guarantee is not sufficient if the uncertainty level increases. A higher uncertainty would increase the divergence possibility and result in a higher task cost.

\section{Differentiable Robust LQG}
We now propose a family of robust LQR-based policies that is based on the robust discrete time-invariant LQG problem in Eq.~\ref{problem}. We consider a differentiable robust LQR program in the infinite-horizon setting. One of the main technical challenges is how to differentiate through such a program with an unbounded dimensionality and uncertain constraints. To tackle this challenge, we will exploit the LMI techniques which are used commonly in optimal control. In particular, rewriting the objective and constraints using LMI techniques would take advantages of robust optimal control and convex optimization. The robust LQR problem's solution can be found by solving a (convex) SDP. This SDP can be solved and differentiated through using the technique developed by Agrawal et. al. \citep{agrawal2019differentiable}, called \emph{differentiable convex optimization layers}. We will discuss how to make this SDP disciplined parameterized programming (DPP) compliant so that it can be differentiated through.

\subsection{Differentiable Infinite-horizon LQR}
\label{diffrobust}
This section starts with the LMI formulation in Eq.~\ref{lmi}, for the robust LQR-based policy whose output is the solution to the LQR problem as written in Eq.~\ref{problem-lqr}. We term this approach as our first contribution, \emph{LMI-LQR layer}, which is considered as an alternative solution for the dynamic Riccati recursion solver used in the differentiable LQR approach described in \ref{difflqr}. The LMI-based LQR layer provides the solution of the SDP program in Eq.~\ref{lmi} as output. This LQR-based policy module is parameterized by parameters $\theta = \{A,B,Q,R\}$. In order for learning with this policy to be end-to-end differentiable, we need an efficient method to compute the derivatives, ${\partial \pi^*}/{\partial \theta}$, of the output policy $\pi^*$ w.r.t the parameters $\theta$. The SDP program in Eq.~\ref{lmi} can be solved using standard convex optimization tools, however it can only be differentiated through efficiently if its constraints and objectives are affine mappings of the problem data (e.g. $\theta$) \citep{agrawal2019differentiating}, i.e. to be DPP-compliant for general \emph{differentiable convex optimization layers} \citep{agrawal2019differentiable}. In this paper, we resort to the \emph{differentiable convex optimization layers} as a differentiation technique, with the introduction of two auxiliary variables $S1, S2$, subject to the additional constraints $S1 = PB, S2=PA$ in order to make the SDP program in Eq.~\ref{lmi} to be DPP-compliant.

The above LMI-based LQR layer cannot directly model uncertainty. Thus this approach will only work well if the uncertainty is small enough. In next section, we will propose a new approach that directly models uncertainty by means of the parameters $\{D,\sigma\}$.





\subsection{Differentiable Robust Infinite-horizon LQR}
\label{diffrobust2}
We now want to explicitly represent the uncertainty and perturbations and learn these terms. As a first step we focus on the dynamics alone by assuming that there is uncertainty about $X=[A,B]$. We denote the uncertainty set as
$$
\Theta =\{X: (X^{\top} -\mu)^\top D(X^{\top} - \mu)\le I\},
$$
where $\mu=[\bar A ,\bar B]$ are the nominal parameters.
Thus, the robust LQR-based policy module has parameters $\theta=\{\bar A ,\bar B,Q,R, D, \sigma\}$. We now discuss how to rewrite our robust problem so that its derivatives w.r.t $\theta$ can be computed efficiently. We follow similar derivations like the system level synthesis framework for ${\cal H}_\infty$ optimal control studied by Anderson et. al. \citep{anderson2019system}, which is recently used in the robust RL framework by Umenberger et. al. \citep{umenberger2019robust}. The robust LQR objective can be rewritten as a worst-case problem, 
\begin{align*}
J(\theta,K) = \min_{K} \mathbb{E} \left[\sup_{\theta\in\Theta }\lim_{T \rightarrow \infty}\sum_{t=0}^T c(x_t,u_t) \right],
\end{align*}
s.t. the linear time-invariant dynamics
$$x_{t+1} = Ax_t + B u_t + w_t, w_t \sim \cc{N}(0,\sigma^2 I), x_0 = x_{\tt{init}}.$$

Assuming the policy parameterization  $u_t=Kx_t$. We can rewrite the infinite horizon cost as
\begin{equation*}
    \small
\begin{aligned}
  &J(\theta,K) =\lim_{T \rightarrow \infty}\frac{1}{T} \expec{E} \left[\sum_{t=0}^T ( x_t^\top Q x_t +u_t^\top R u_t) \right] \\&= \tt{trace}
  \left( \left[\begin{array}{cc} Q & 0 \\ 0 &R \end{array} \right] \lim_{T \rightarrow \infty}\frac{1}{T}\sum_{t=1}^T \expec{E}
\left[ \left[ \begin{array}{c} x_t \\ u_t\end{array} \right]^\top \left[ \begin{array}{c} x_t \\ u_t\end{array} \right]  \right] 
\right) \\
& =\tt{trace}
  \left( \left[\begin{array}{cc} Q & 0 \\ 0 &R \end{array} \right] \lim_{T \rightarrow \infty}\frac{1}{T}\sum_{t=1}^T \expec{E}
\left[ \left[ \begin{array}{c} x_t \\ Kx_t \end{array} \right]^\top \left[ \begin{array}{c} x_t \\ Kx_t \end{array} \right]  \right]  
\right) \\
& = \tt{trace}
  \left( \left[\begin{array}{cc} Q & 0 \\ 0 &R \end{array} \right] \left[ \begin{array}{cc}W & WK^{\top} \\ KW & KW K^{\top} \end{array}\right] \right),
\end{aligned}
\end{equation*}
where we denote $W = \expec{E}\left[x_t x_t^{\top}\right]$, the stationary state covariance. Given certain $A$ and $B$, $W$ can be found by solving the following optimization problem,
\begin{equation}
\small
\begin{aligned}
  &\argmin_W \tt{trace} (W), \\
  & \text{s.t.} \quad W  \succeq (A+BK) W (A+BK) ^{\top}  + \sigma^2 I .
\end{aligned}
\label{eqminW}
\end{equation}
If we assume that the policy has only parameters $K$, we receive a semidefinite program (SDP), which is convex. However, in the case of uncertain $\{A,B\}$, solving the above program is very challenging \citep{el1998robust}.

We will now reformulate it by resorting to a worst-case scenario. Using similar notations used as by Umenberger et. al. \citep{umenberger2019robust}. We denote
$Z = WK^{\top}, Y = KW K^{\top} $, and
\begin{align}
  \Xi = \left[ \begin{array}{cc} W & Z \\ Z ^{\top} & Y \end{array} \right] .
  \label{def:xi}
\end{align}
The non-convex semidefinite constraint in Eq.~\ref{eqminW} can be rewritten as a convex one as
\begin{align} 
  \quad W  \succeq X \Xi X ^ {\top} + \sigma^2 I ,
  \label{cons:sdp}
\end{align}
where according to the the Schur complement, it can be further rewritten as 
\begin{align*}
\left[\begin{array}{cc}
     I& \sigma I  \\
     \sigma I & W - X \Xi X^{\top}
\end{array} \right ] \succeq 0 .
\end{align*}
The second challenge of the uncertain parameters can be handled through the use of the worst-case formulation as
\begin{equation}
  \begin{aligned}
    \min_W  \tt{trace} &
    \left( \left[\begin{array}{cc} Q & 0 \\ 0 &R \end{array} \right] \Xi \right) \\
    \text{s.t.}  \quad & \left[\begin{array}{cc}
     I& \sigma I  \\
     \sigma I & W - X \Xi X^{\top}
\end{array} \right ] \succeq 0, \\ & (X^{\top} -\mu)^{\top} D (X^{\top} -\mu) \le I,
  \end{aligned}
  \label{eq:upp}
\end{equation}
where $\Xi$ is defined in Eq.~\ref{def:xi}. In order to further rewrite the uncertain constraints in Eq.~\ref{eq:upp}, we follow similar derivations from \citep{umenberger2019robust,luo2004multivariate}. We use the following Theorem from \citep{luo2004multivariate} (Theorem 3.7).
\begin{theorem} The data matrices $({\cal A},{\cal B},{\cal C},{\cal D},{\cal F},{\cal G},{\cal H})$ satisfy the robust fractional  quadratic matrix inequalities
\begin{align*}
&\left[\begin{array}{cc}
     {\cal H}& {\cal F + GX} \\
     {\cal(F+GX)}^{\top}& {\cal C}+ {\cal X}^\top{\cal B + B}^{\top} {\cal X} + {\cal X}^{\top} {\cal A X}
\end{array} \right]\succeq 0 ,\\
& \forall {\cal X} \quad \text{such that} \quad I- {\cal X}^{\top}{\cal D}{\cal X} \succeq 0, 
\end{align*}
if and only if there is $\lambda \ge 0$ such that
\begin{align*}
&\left[\begin{array}{ccc}
     {\cal H}& {\cal F} & {\cal G} \\
     {\cal F}^{\top}& {\cal C} - \lambda  I& {\cal B}^{\top} \\
     {\cal G}^{\top}  & {\cal B} & {\cal A} + \lambda {\cal D}
\end{array} \right]\succeq 0 .
\end{align*}
\end{theorem}
We use substitutions similar to the one in \citep{umenberger2019robust}: ${\cal A}=-\Xi$, ${\cal B}=\Xi [\bar A, \bar B]^\top$, ${\cal C}=W-[\bar A, \bar B] \Xi [\bar A, \bar B]^\top$, ${\cal D}=D$, ${\cal F}=\sigma I$, ${\cal G}=0$, ${\cal H}=I$. In addition, we substitute ${\cal X} = X - \mu $ with the definition of $X=[A,B] = [\bar A, \bar B] -\cal X$. 

As a final result of the substitutions, we receive the program in Eq.~\ref{eq:upp} as a SDP,
\begin{equation}
\small
  \begin{aligned}
&\min_{\Xi,\lambda}  \tt{trace} 
    \left( \left[\begin{array}{cc} Q & 0 \\ 0 &R \end{array} \right] \Xi \right) \\
     &  \text{s.t.} \quad   \left[\begin{array}{ccc} I & \sigma I & 0 \\
        \sigma I & W - [\bar A\quad \bar B] \Xi [\bar A\quad \bar B]^{\top} & [\bar A\quad \bar B] \Xi^{\top} \\
        0 & \Xi [\bar A\quad \bar B]^{\top} & \lambda D - \Xi
      \end{array}\right]   \succeq 0, \\ & \lambda \ge 0.
  \end{aligned}
  \label{eq:sdp}
\end{equation}

Solving the above cone program, we receive the optimal $W$, that helps to reconstruct the policy as $K=Z^{\top}W^{-1}$. We are interested in computing the derivative:
\begin{align}
  \frac{  \partial l}{\partial \theta} = \frac{  \partial l}{\partial K}  \frac{  \partial K}{\partial W}\frac{\partial W}{\partial \theta}
  \label{eq:chainrule}
\end{align}
where $\theta=[\bar A, \bar B, Q,R,D, \sigma]$ are the robust LQR's parameters. The scalar function $l$ is the task objective depending on the policy (parameterized by $K$), i.e. imitation learning cost. The middle part in R.H.S can be computed easily, according to the definition of the resulting policy where $W$ is the optimal solution of the SDP problem. The last derivative can be computed by differentiating through a cone program, for which we can utilize the general approach proposed in a recent work by Agrawal et. al. \citep{agrawal2019differentiating}, differentiating through a cone program. Our practical implementation needs the introduction of an auxiliary variable $S$, with an additional constraint $S=\Xi [\bar A,\bar B]$. An extension to modeling also the uncertainty $Q$ and $R$ is simple, which has little intuition so we do not consider in this work. 



\section{Experiments}
In this section, we evaluate the differentiable robust LQR layers, i.e. the LMI-based LQR (LMI-LQR) and LMI-based robust LQR (LMI-Robust-LQR) layers proposed in \ref{diffrobust} and \ref{diffrobust2}, in terms of their performance and uncertainty handling capabilities in comparisons to other baseline methods. Our main contender is the differentiable LQR framework (mpc.pytorch) \citep{AmosRSBK18}. Both LMI-LQR and mpc.pytorch can be considered as a solver for the nominal system. Our implementation is based on PyTorch and uses cvxpylayers \citep{agrawal2019differentiable} as the main differentiable solver for both LMI-LQR and LMI-Robust-LQR layers (where a SCS cone programming solver \citep{ocpb:16} is chosen by default).


\subsection{Imitation Learning on Robust LQR}
In this section we design a robust LQR task with dynamics as defined in Eq.~\ref{problem}. The expert is parameterized with a robust LQR controller. This assumption also reflects the reality since under many uncertain scenarios in nature human behave following to some worst-case strategies \citep{lipshitz1997coping}. Three learners are mpc.pytorch, LMI-LQR, and LMI-Robust-LQR with different parameterization. We assume that the quadratic costs $Q=R=I$, state and control dimensions $n=3,m=3$ and the variance of the noise $\sigma=0.1$ are known to the learners. The learners are supposed to learn the the parameters $\bar A$ and $\bar B$ of the linear system dynamics and the model uncertainty $D$. Note that mpc.pytorch and LMI-LQR can not learn $D$. The experts are generated from random $\bar A,\bar B,D$ (where we additionally control their stability), depending on a random seed. All algorithms use the same initialization for $\bar A$, $\bar B$, $D$. All algorithms are implemented using PyTorch with the following settings: RMSprop optimizer (momentum $= 0.5$, learning rate $= 0.01$), a minibatch of $16$ trajectories. Given demonstration data ${\cal D}=\{\tau^*_i\}_{i=1}^N$, all algorithms use a similar \emph{imitation objective} for training, $
l =\frac{1}{\|\cal D\|}\sum _i \| \tau^*_i -\tau_i\|^2 
$. For validation, we generate separate $N=32$ trajectories $\tau^*_i$ using the expert's robust LQR policy. Each generated trajectory $\tau_i$ is conditioned on only an initial state $x^*_0$ that initiated the expert trajectory $\tau^*_i$. The \emph{model loss} is defined as a $l_2$ norm of the difference between the estimate vs. the ground-truth: e.g. $\bar A,\bar B$ vs. $\bar A^*,\bar B^*$ (scenario 1), or $D$ vs. $D^*$ (scenario 2). This loss is used only for reporting, which measures how the model-based module works based purely on a model-free loss (the imitation loss). The \emph{validation cost} is obtained by running optimized controllers on the true stochastic dynamics and a known uncertainty set.

All results are averaged over ten random seeds. Experiments are run on an Intel i7 CPU (2.6Ghz, 12 core).




\paragraph{Scenario 1: Known $\bar A,\bar B$, unknown model uncertainty $D$}
In this scenario only $D$ has to be learned. Since the differentiable LQR framework \citep{AmosRSBK18} and LMI-based LQR layer do not have uncertainty modeling, they do not require training. On the other hand, we train LMI-based robust LQR for 200 iterations to optimize the imitation loss w.r.t parameter $D$, where we assume both the true $D^*$ and $D$ are diagonal. We first evaluate the solving time of different algorithms. Table \ref{table1} shows the total computation time (forward passes) of three algorithms on different horizon lengths. The results show that our differentiable infinite-horizon (robust) LQR layer is more computationally efficient by a factor of the horizon length. 

\begin{figure}
  \includegraphics[width=0.245\textwidth]{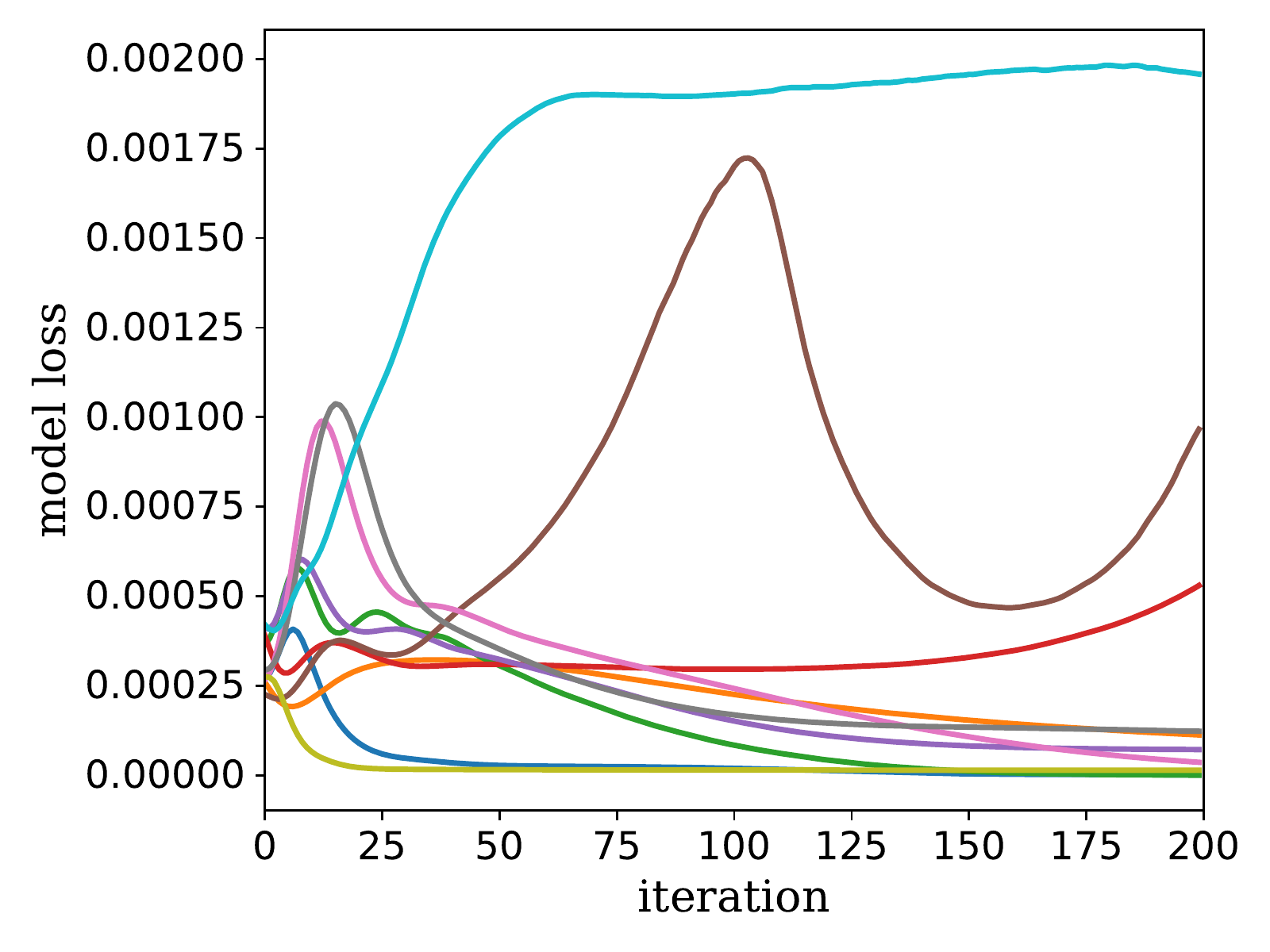}\includegraphics[width=0.245\textwidth]{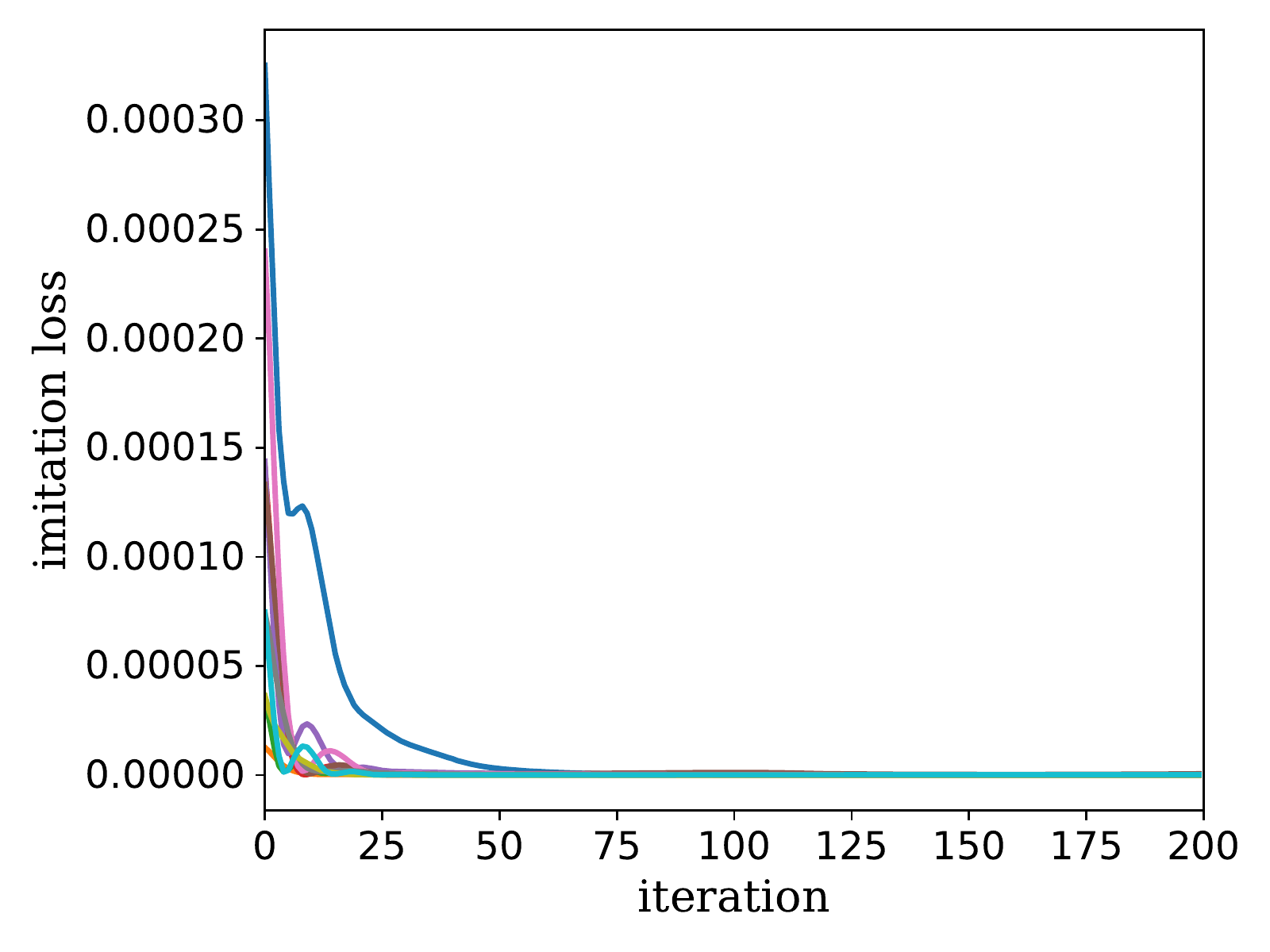}
  \caption{Model and imitation losses for 10 different runs (each plot corresponds to one run).}
  \label{fig:uncertainty-loss}
\end{figure}

Table \ref{table2} (1st row: S1) shows the performance comparison of optimal controllers found by the three algorithms in terms of the validation cost. Only the LMI-based robust LQR layer algorithm requires training, so the performance of the final controller at convergence is reported. This result shows that mpc.pytorch performs poorly because it is not optimizing a controller that can be robust under uncertainty. On contrary, the differentiable LMI-based LQR layer method which incorporates robust control constraints via a LMI formulation performs much better. This shows the benefit of using a robust control constraint to stabilize the optimized controller as output. As this simple approach does not model uncertainty directly, therefore it can only stabilize the output controller within small bounded perturbations. In this experiment, the environment uncertainty is set to a high value, therefore we can see in Table \ref{table2} (1st row: S1) the performance of LMI-LQR is not optimal. This drawback is addressed by the LMI-based robust LQR layer method where its validation cost is significantly better than mpc.pytorch and LMI-LQR. This performance level is equal to the optimal cost received by running an optimal worse-case policy found on the true model. The plots of the model and imitation losses are reported in Figure \ref{fig:uncertainty-loss}. Similar to findings in mpc.pytorch \citep{AmosRSBK18}, the imitation loss might converge to a local optima, while there are possible divergences of the the model loss. This shows the challenges of optimizing a highly non-linear layer in which its weights are from parameters of an optimal control program.

\begin{table}
  \caption{Running time (in seconds) for solving a batch of 128 sampled LQR problems, with a varying length of horizon.}
  \label{table1}
\begin{center}
\begin{tabular}{|c|c|c|c| } 
  \hline
 Time horizon & 10 & 50 &100 \\
  \hline
 mpc.pytorch &15 & 71.9 & 139.9 \\
 LMI-LQR &0.99 & 0.99  &0.99  \\ 
 LMI-Robust-LQR &1.67& 1.67&  1.67  \\ 
 \hline
\end{tabular}
\end{center}
\end{table}

\begin{figure*}
\center
  \includegraphics[width=0.3\textwidth]{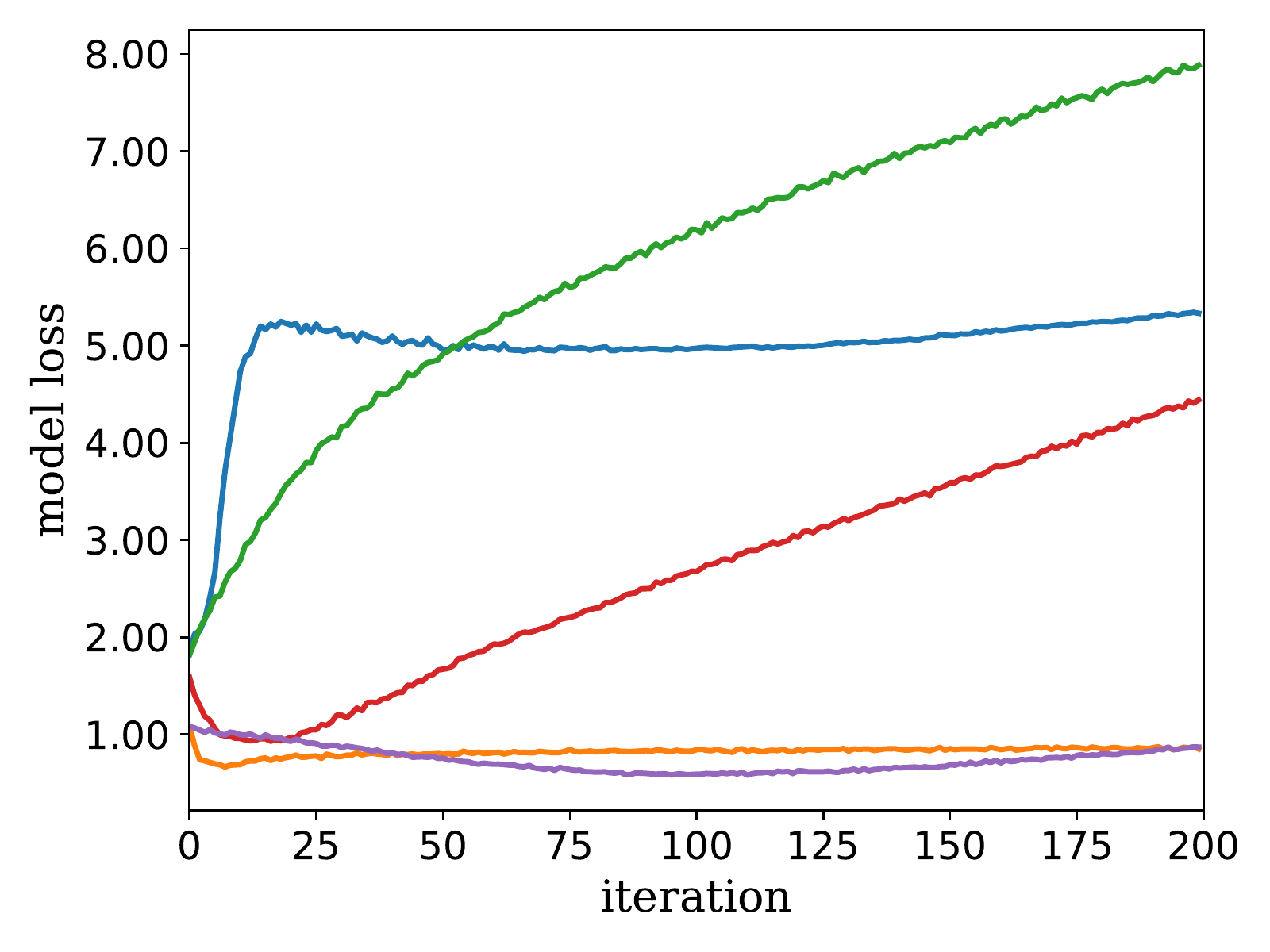}\includegraphics[width=0.3\textwidth]{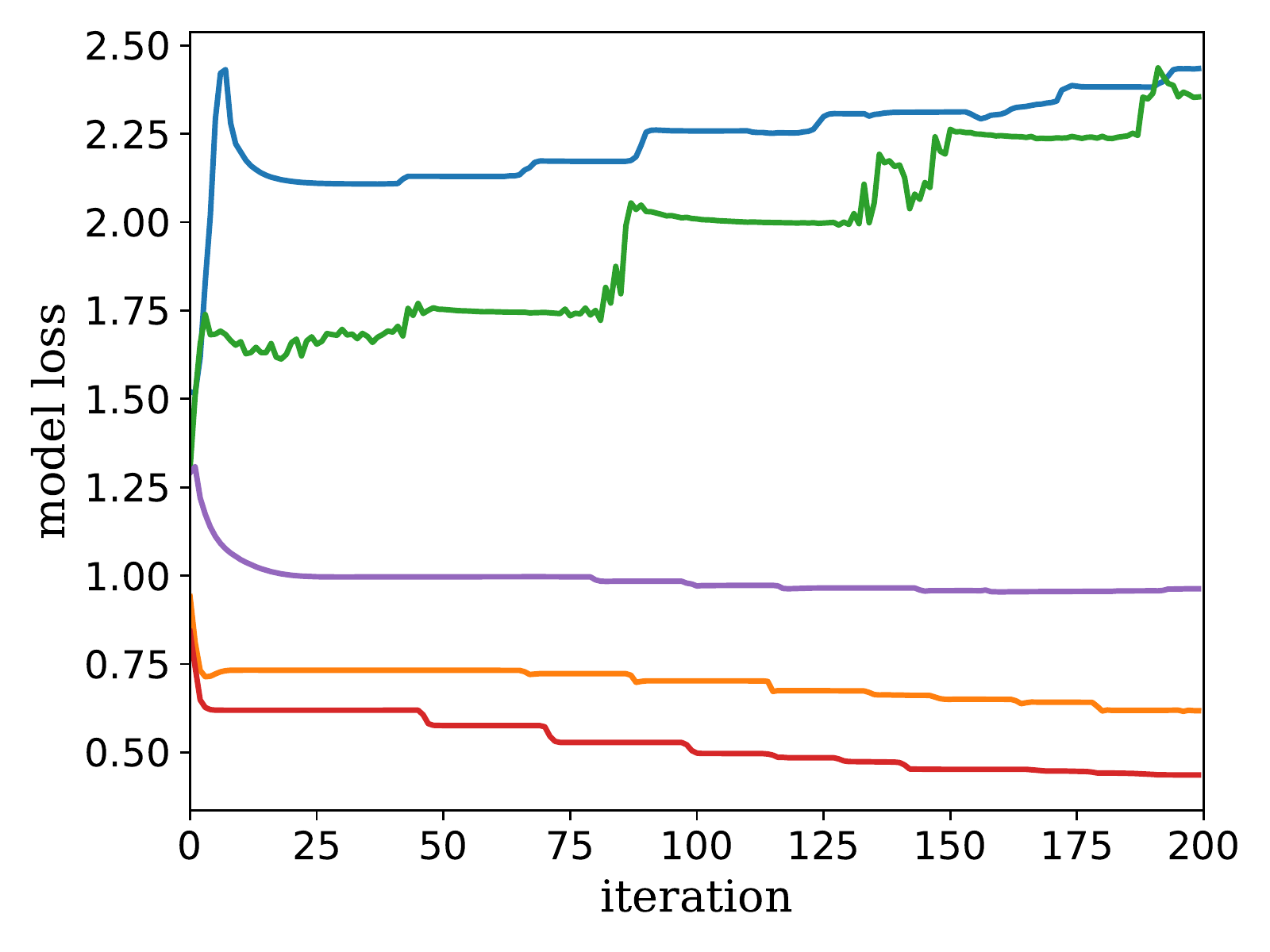}\includegraphics[width=0.3\textwidth]{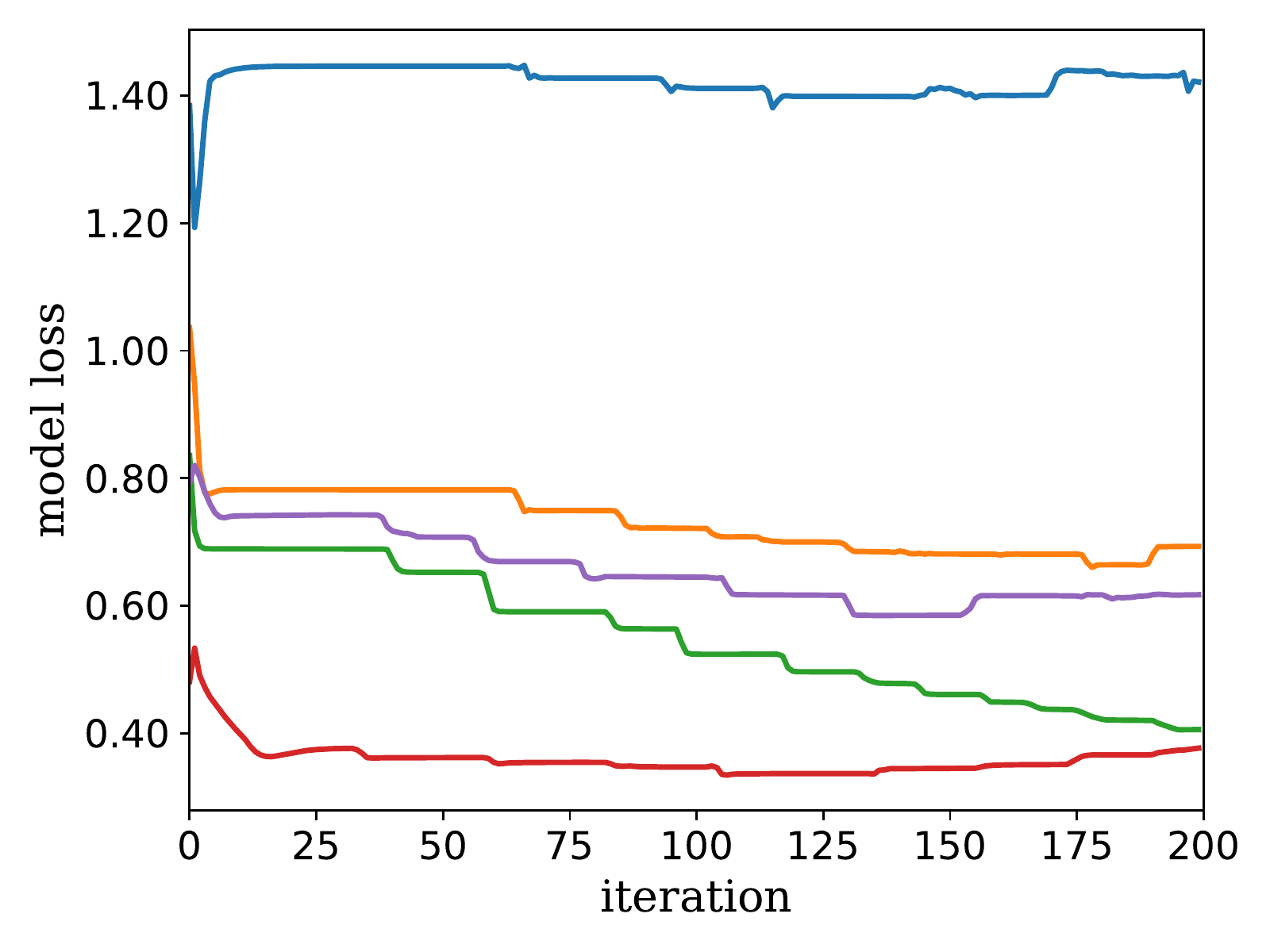}
  \includegraphics[width=0.3\textwidth]{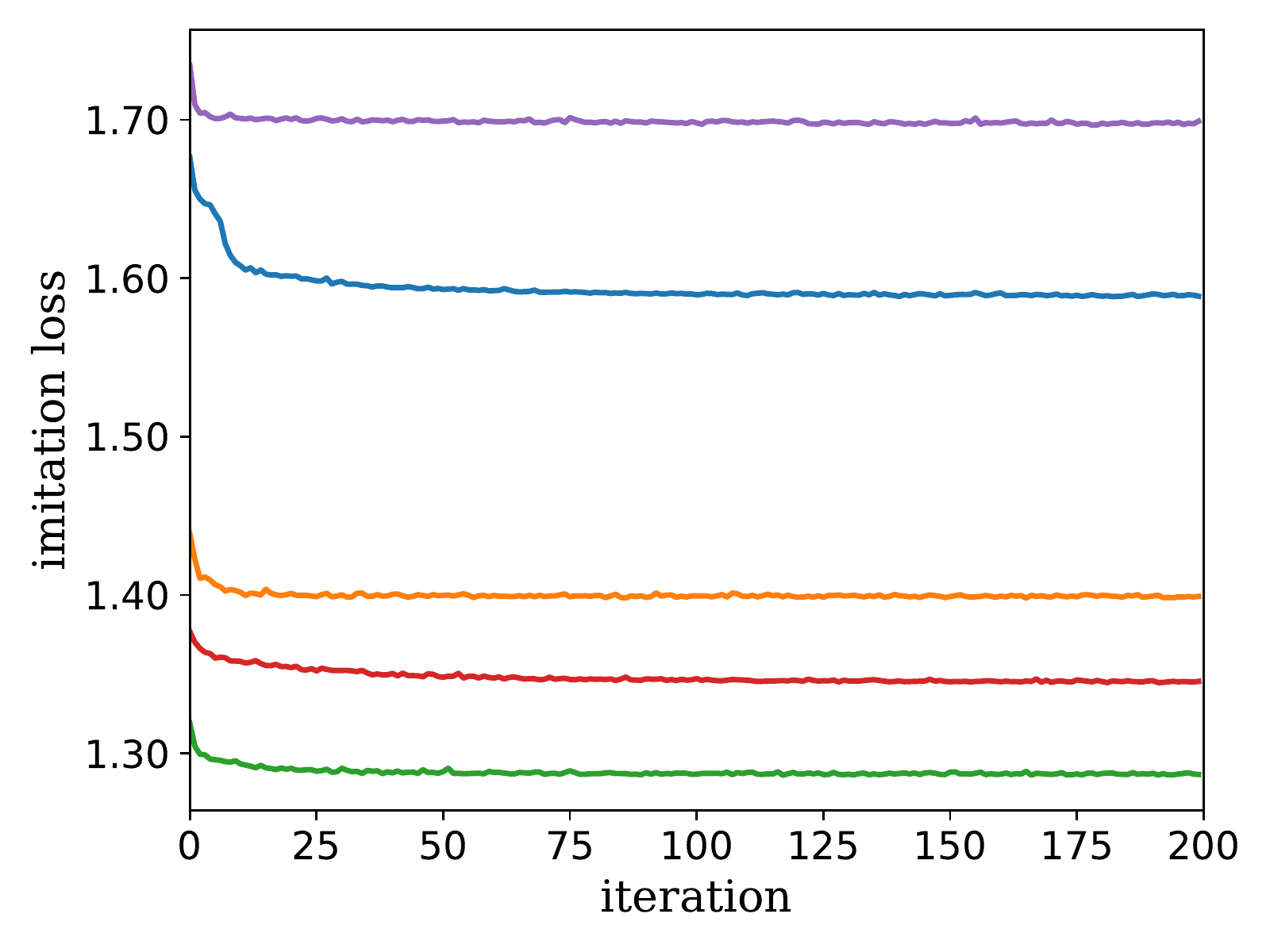}\includegraphics[width=0.3\textwidth]{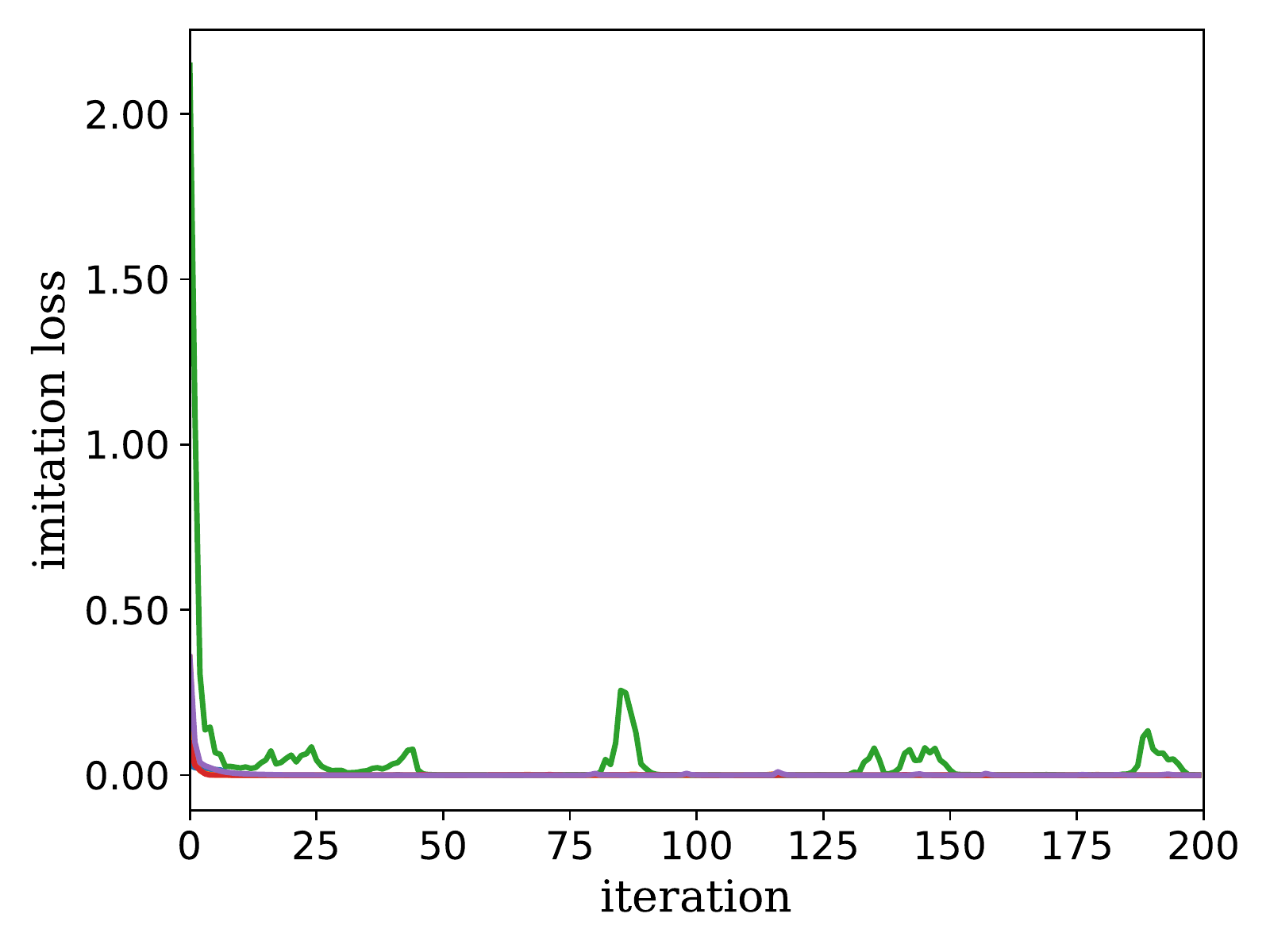}\includegraphics[width=0.3\textwidth]{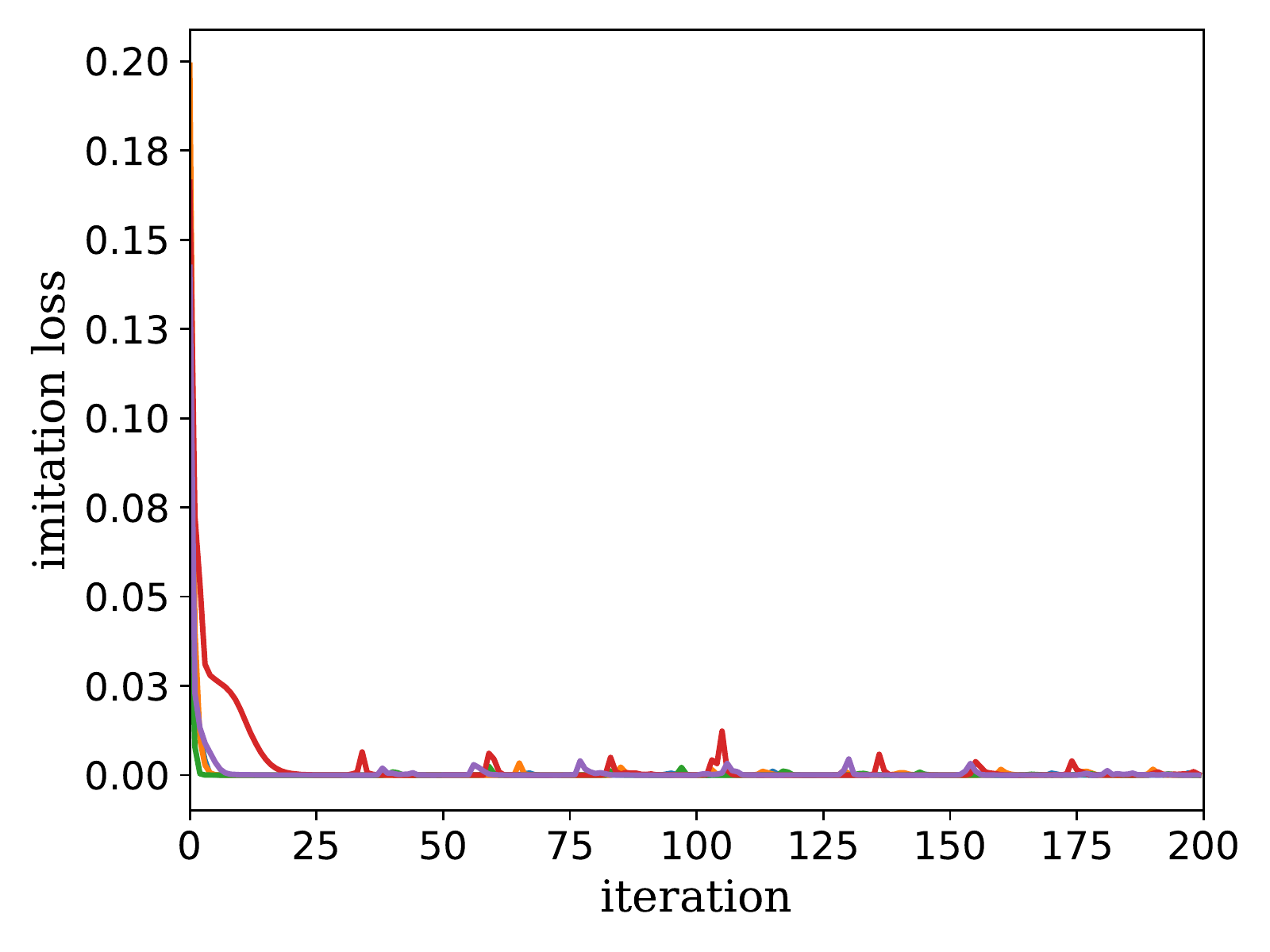}
  \caption{Model and imitation losses (5 runs) for: mpc.pytorch (left), LMI-LQR (middle), LMI-Robust-LQR (right).}
  \label{fig:scenario2}
\end{figure*}

\paragraph{Scenario 2: Known uncertainty $D$, unknown model $\bar A,\bar B$}
In this scenario,we evaluate the performance of the different algorithms on imitation learning, where $D$ is known and $\bar A,\bar B$ are supposed to be learnt. All algorithms are initialized randomly with the same stable $\bar A,\bar B$, with a horizon length of 20. Figure \ref{fig:scenario2} shows the imitation and model losses of mpc.pytorch (right), LMI-LQR (middle), LMI-Robust-LQR (left). While the model losses may not be suitable to assess the performance of three differentiable layers (a similar observation was made in \citep{AmosRSBK18}), the imitation losses can reflect the actual quality of the controls, which are generated by these layers. The results show that mpc.pytorch converges only to a local optima that still has a large imitation loss. While LMI-LQR can achieve an optimal imitation loss, its optimized controller is not robust enough as reflected by its validation cost in Table \ref{table1} (2nd row: S1). LMI-Robust-LQR leads to controller that is more robust because it has an ability to learn model uncertainties.

\begin{table}
  \caption{Validation cost evaluated on the true dynamics, with a horizon set to 50. Both the mean cost and its standard deviation over 10 different runs are reported.}
  \label{table2}
\begin{center}
  \begin{tabular}{|c|c|c|c| }
    \hline
   &  mpc.pytorch & LMI-LQR & LMI-Robust-LQR \\
    \hline
    S1&  $44.1 \pm 1.6$ &  $41.2 \pm 1.4$  & $\bf 10.9 \pm 0.5$   \\
    \hline
  S2 &$124.5 \pm 10.2 $ &  $67.4 \pm 18.7  $  & $\bf 11.8 \pm 2.3$   \\
  \hline
\end{tabular}
\end{center}
\end{table}

\subsection{Convex Approximate Dynamic Programming}
In this section we evaluate our methods with full functionalities, i.e. we differentiate and learn all parameters: $\{A,B,Q,R\}$ in LMI-LQR layer and $\{\bar A, \bar B,Q,R,D,\sigma\}$ in LMI-Robust-LQR. For this purpose, we evaluate the proposed algorithms on an uncertain stochastic optimal control (SOC) problem,
\begin{equation}
\small
\begin{aligned}
  &\text{minimize}_{u=\pi(x)} \quad {\tt{lim}}_{T\rightarrow \infty} \mathbb{E}\left[ \frac{1}{T}\sum_{t=0}^{T-1}\|x_t\|^2_2 +\|u_t\|_2^2 \right] \\
   \text{s.t.} \quad& x_{t+1} =Ax_t + Bu_t + w_t \\  \quad & w_t \in {\cal N}(0,\sigma^2I_n) ,\quad x_0 \sim P_0,\\
  & [A,B]  \in \{A,B: \left(X^{\top} - \mu \right) ^{\top} D \left(X^{\top} - \mu \right) \le I\}
\end{aligned}
\label{socproblem}
\end{equation}
where $X$ denotes $[A,B]$, and $\mu=[\bar A, \bar B ]$. This problem introduces an uncertainty set $D$ over model parameters $A,B$, which is different from a similar problem for a nominal dynamic system considered by Agrawal et. al. \citep{agrawal2019differentiable}. We evaluate three different policy parameterization that are all based on the Lyapunov stability theory \citep{boyd1994linear}, and three simple baselines that do not have optimal control base. 

\paragraph{Simple baselines} We use three baseline methods that do not use differentiable optimal control layers, denoted as $u=f_\theta(x)$: (1) a linear controller $u=Kx$, where $\theta=K$ is a parameterized \emph{feedback gain} matrix; 2) $f_\theta$ is a multi-layer perceptron (MLP) with two hidden layers of shape $(32,32)$ using ReLU activations; and 3) $f_\theta$ is a long-short term memory network (LSTM) (a recurrent controller) with one 64-unit hidden layer. All three baselines use the Adam optimizer with a step-size $\alpha=1e-4$. 

\begin{figure*}[t]
  \includegraphics[width=0.33\textwidth]{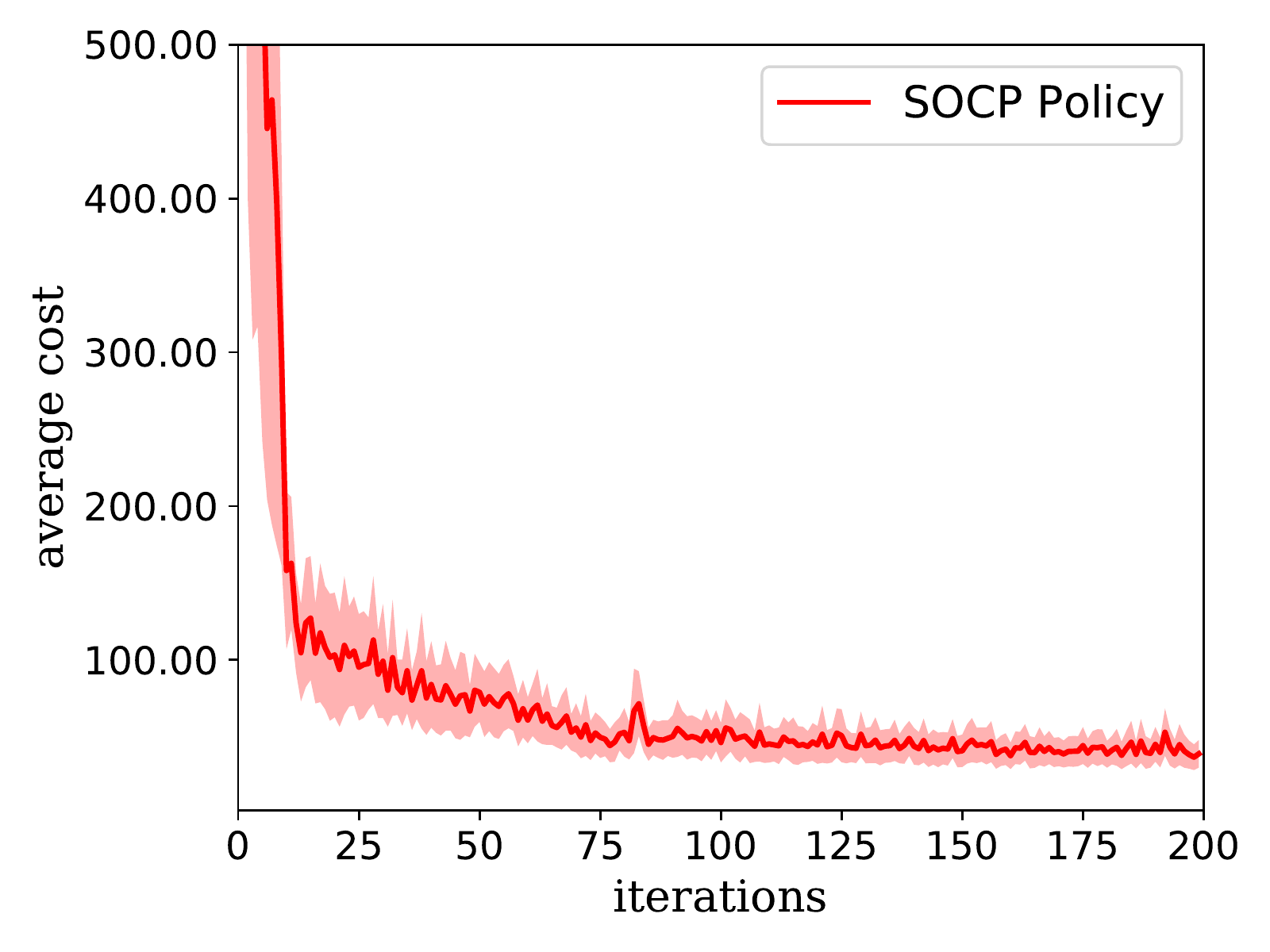}\includegraphics[width=0.33\textwidth]{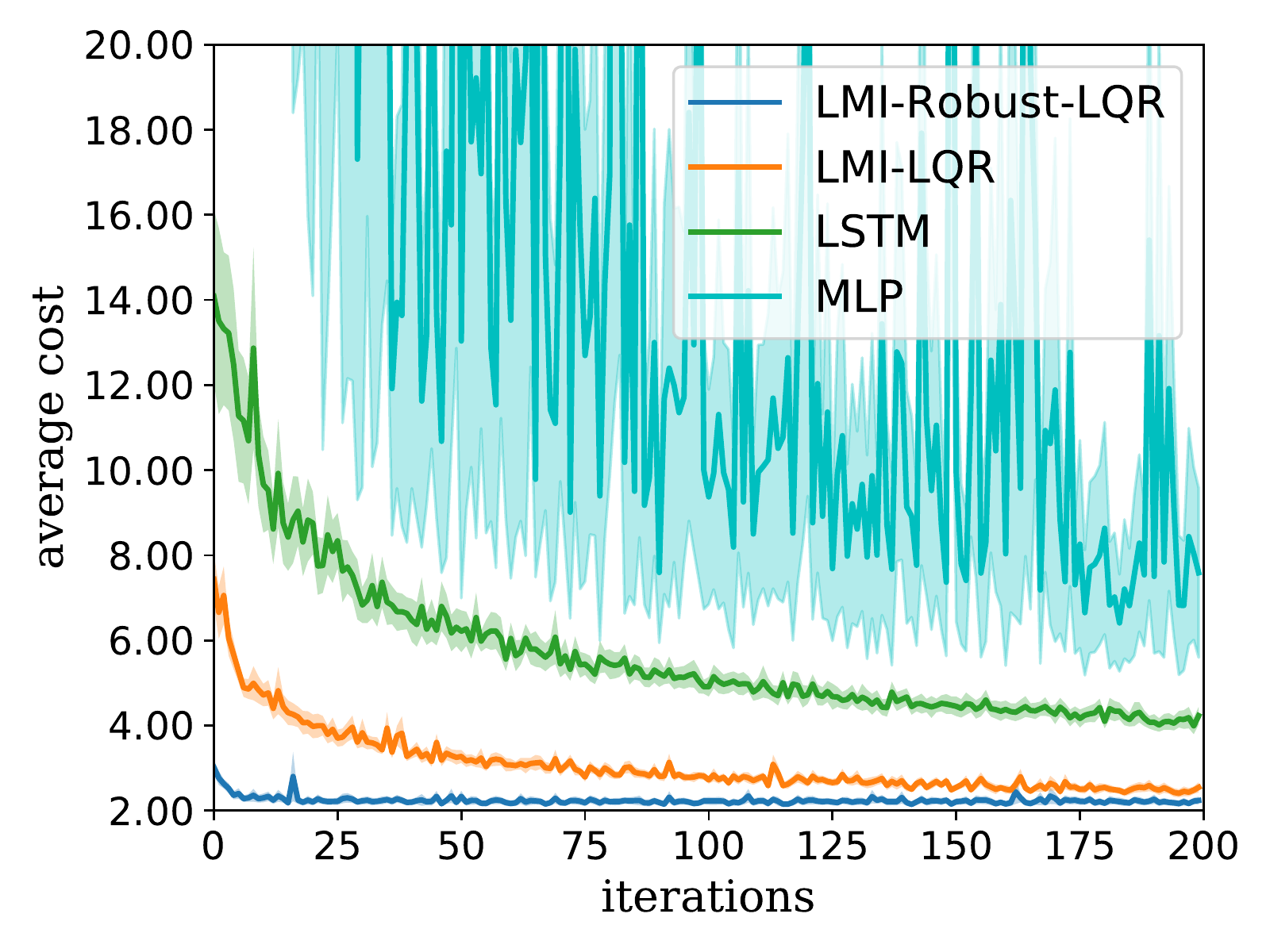} \includegraphics[width=0.33\textwidth]{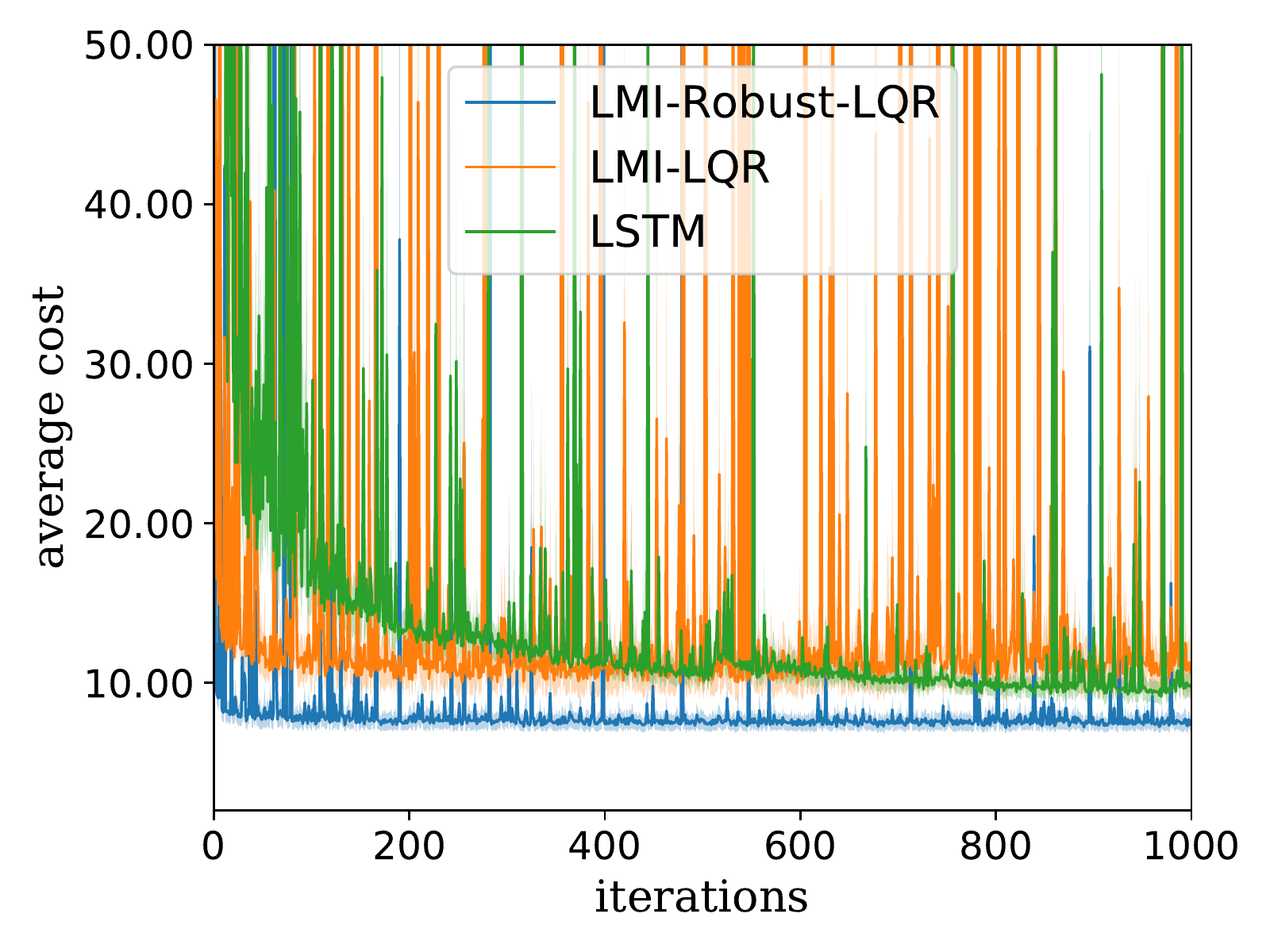}
  \caption{Average cost for robust SOC with a horizon of 20 (left and center), and 100 (right). The plot of the SOCP policy is separated due to its large domain. The shading is the standard deviation of the mean estimator.}
  \label{fig:adp-cost}
\end{figure*}

\paragraph{Optimal control-based policies} Besides our two proposals LMI-LQR, LMI-Robust-LQR, the next baseline is an approximate dynamic programming policy (ADP) from \citep{agrawal2019differentiable}. They proposed to use a quadratic control-Lyapunov policy as output of a differentiable convex layer. This convex layer is designed to solve the following second-order cone program (SOCP) subject to a constraint of bounded controls,
\begin{equation}
\begin{aligned}
  \text{minimize}_{u} \quad & u^{\top} P u + x_t^{\top}Q u + q^{\top} u \\
   \text{s.t.} \quad& \|u\| \le 1
\end{aligned}
\label{socppolicy}
\end{equation}
This SOCP policy receives $u$ as variable, and $\{P,Q,q,x_t\}$ as parameters. Using a differentiable convex layer \citep{agrawal2019differentiable}, we can differentiate through the program in Eq.~\ref{socppolicy}. As an alternative approach, our two proposed differentiable LQR layers are based on the Lyapunov inequality where we parameterize the policy directly, $\pi$ is the optimal solution of the program defined in Eq.~\ref{problem}. Our differentiable layers receive $K$ (as defined in Section \ref{diffrobust}) as variable (where $T$ is the horizon length), and $\{A,B,Q,R\}$ (for LMI-LQR layer) and $\{\bar A, \bar B,Q,R,D,\sigma\}$ (for LMI-Robust-LQR) as parameters. Note that both the LMI-LQR layer and the SOCP policy cannot model the uncertainty set directly, hence they are just solvers for a nominal system (without uncertainty). 

We use state and action dimensions $n=m=3$, a time horizon of $T=20$, a batch size of 64 randomly initial states, and run for 200 update iterations. We randomize different SOC problems in Eq.~\ref{socproblem}: the dynamics model $\bar A,\bar B$ are initialized with Lyapunov-stable matrices, a random diagonal matrix $D$ with diagonal entries, and a fixed noise variance $\sigma=0.1$. All algorithms use the same RMSprop optimizer with their optimized learning rate and decay rate, i.e. using grid-search. The parameters of the SOCP policy are initialized as in the original paper of Agrawal et. al. \citep{agrawal2019differentiable} where $\{P,Q,q\}$ are initialized with the exact LQR solution of the program in Eq.~\ref{socppolicy} without constraints on actions $u$. We initialize the parameters $\{\bar A,\bar B,Q,R,D,\sigma\}$ randomly (using Gaussian distributions) for the  LMI-LQR and LMI-Robust-LQR layers. The evaluations are averaged over 10 different random seeds. We report the mean cost and its standard deviation.

\paragraph{Results} Figure \ref{fig:adp-cost} reports the average cost optimized by each algorithm. While the first simple baseline diverges on both tasks with a horizon of 20 and 100, hence they are not reported in the plots. Its evaluated cost (the objective in Eq. 12) always returns $\mathrm{nan}$ or $\infty$. The second baseline using MLP may solve the task with a horizon of 20, but shows a very unstable behavior due to fluctuated cost evaluations, i.e. spikes as seen on the plots. As a result, it is not surprised that it becomes diverging when the horizon increases to 100 (we also design this long horizon task with a higher dynamics stochasticity $\sigma=0.5$), hence not reported in the (right) plot. The third baseline using LSTM is known to learn better for tasks under uncertainty, e.g. POMDP. However on this challenging robust control task, it performs worse than our proposed methods in terms of sample-efficiency and the final task cost. The main reason is that LSTM is a general neural network that can not exploit the inherent structure of the task, i.e. uncertainty constraints like in our method. This is further demonstrated on the right plot for the task with a horizon of 100 in which the LSTM controller is very unstable (in this task we set a longer horizon and a higher variance for the stochastic dynamics).

The results also show that directly incorporating robust constraints in LMI-LQR and LMI-Robust-LQR in the layer, i.e. via the Lyapunov inequality, helps achieve much lower cost. The SOCP policy can only converge to an inferior local policy that incurs a very high cost. A direct principled uncertainty modeling like in LMI-Robust-LQR achieves the best overall performance. In addition, we observe that the total training time of one trial for each algorithm on a personal workstation with a modern CPU is 1-2 hours for SCOP policy, 20-30 minutes for LMI-LQR, and 30-45 minutes for LMI-Robust-LQR. This running time increases approximately linearly with respect to the time horizon.

As the time horizon increases, the effect of uncertain model and stochastic dynamics can lead the system to instability. We rerun all three algorithms for 1000 training iterations. All trials using the SOCP policy return numerical errors within few iterations (so no plots are reported for the SOCP policy). The validation costs for LMI-LQR and LMI-Robust-LQR are reported in Fig.~\ref{fig:adp-cost} (right). The results show that there is an extreme effect of uncertainty and stochasticity to training. While LMI-LQR seems to gradually attenuate the effect of perturbations, it might take the training very long to achieve a stable result. In contrary, LMI-Robust-LQR reduces the effect of uncertainty well. Its final controller is less fluctuating and starts becoming stable.


\section{Conclusion}
This paper proposes a new differentiable optimal control-based layer. Our approach is motivated by recent work on using convex optimization inside neural network layers. We directly use a robust optimal control optimization program as a differentiable layer that can be incorporated in standard end-to-end neural networks. Our main contribution, which differentiates our work from the current state-of-the-art, is to model uncertainty and stochasticity directly inside a differentiable optimal control layer. As a result, our proposed optimal control layers can optimize a controller as output that is stable and robust to perturbations and noises residing in demonstration data. Our layer can be integrated as a differentiable controller in reinforcement learning and learning from demonstration on tasks where uncertainty and perturbations are present.

Using an infinite-horizon LQR setting has helped accelerate the computation time of both forward and backward passes significantly. This important achievement would open a variety of more applications for the differentiable LQR/MPC layers, for example on practical robotics tasks or hybrid model-based and model-free policy optimization. One main drawback of this setting is that the output is a linear controller that will have limitations on highly non-linear control tasks. One potential research direction can look at extending to robust time-dependent and finite-horizon settings for either LQR or MPC control theory.

\begin{acknowledgements} 
    Briefly acknowledge people and organizations here.

    \emph{All} acknowledgements go in this section.
\end{acknowledgements}

\bibliography{bib}

\appendix

\end{document}